\documentclass{article} 
\usepackage{iclr2023_conference,times}

\usepackage{hyperref}
\usepackage{url}

\usepackage{mathptmx}
\usepackage[T1]{fontenc}

\usepackage{amsmath,amsfonts,bm}

\def\eqref#1{equation~\ref{#1}}

\def\1{\bm{1}}

\DeclareMathAlphabet{\mathsfit}{\encodingdefault}{\sfdefault}{m}{sl}
\SetMathAlphabet{\mathsfit}{bold}{\encodingdefault}{\sfdefault}{bx}{n}

\usepackage{graphicx}
\usepackage{subfig,xspace}
\usepackage{wrapfig}
\usepackage{todonotes}
\usepackage{booktabs}
\usepackage{makecell}
\usepackage{multirow}

\usepackage{xspace}
\usepackage{units}

\newcommand{\cgpt}{{\sc C-GPT}\xspace}
\newcommand{\cbert}{{\sc C-BERT}\xspace}
\newcommand{\cmtm}{{\sc C-MTM}\xspace}
\newcommand{\csmart}{{\sc C-SMART}\xspace}
\newcommand{\mtm}{{\sc MTM}\xspace}
\newcommand{\smart}{{\sc SMART}\xspace}
\newcommand{\bert}{{\sc BERT}\xspace}
\newcommand{\gpt}{{\sc GPT}\xspace}
\newcommand{\gato}{{\sc GATO}\xspace}

\newcommand{\pasta}{{\sc PASTA}\xspace}

\newcommand{\ie}{\textit{i.e.}\xspace}
\newcommand{\eg}{\textit{e.g.}\xspace}

\definecolor{myred}{rgb}{0.8,0,0}
\definecolor{mygreen}{rgb}{0,0.6,0}
\definecolor{myblue}{rgb}{0,0,0.7}
\definecolor{stanfordblue}{HTML}{006eb8}

\hypersetup{
  colorlinks   = true, 
  urlcolor     = stanfordblue, 
  linkcolor    = stanfordblue, 
  citecolor   = stanfordblue 
}

\title{PASTA: Pretrained Action-State Transformer Agents}

\author{Raphael Boige\thanks{Equal contribution.}, Yannis Flet-Berliac\footnotemark[1], Arthur Flajolet, Guillaume Richard, Thomas Pierrot\\
InstaDeep\\
\texttt{\{r.boige,y.flet-berliac\}@instadeep.com} \\
}

\iclrfinalcopy 
\begin{document}

\maketitle

\begin{abstract}
  Self-supervised learning has brought about a revolutionary paradigm shift in various computing domains, including NLP, vision, and biology. Recent approaches involve pre-training transformer models on vast amounts of unlabeled data, serving as a starting point for efficiently solving downstream tasks. In reinforcement learning, researchers have recently adapted these approaches, developing models pre-trained on expert trajectories. This advancement enables the models to tackle a broad spectrum of tasks, ranging from robotics to recommendation systems. However, existing methods mostly rely on intricate pre-training objectives tailored to specific downstream applications. This paper conducts a comprehensive investigation of models, referred to as \underline{p}re-trained \underline{a}ction-\underline{s}tate \underline{t}ransformer \underline{a}gents (\pasta). Our study covers a unified methodology and covers an extensive set of general downstream tasks including behavioral cloning, offline RL, sensor failure robustness, and dynamics change adaptation. Our objective is to systematically compare various design choices and offer valuable insights that will aid practitioners in developing robust models. Key highlights of our study include tokenization at the component level for actions and states, the use of fundamental pre-training objectives such as next token prediction or masked language modeling, simultaneous training of models across multiple domains, and the application of various fine-tuning strategies. In this study, the developed models contain fewer than 7 million parameters allowing a broad community to use these models and reproduce our experiments. We hope that this study will encourage further research into the use of transformers with first principle design choices to represent RL trajectories and contribute to robust policy learning.
\end{abstract}

\section{Introduction}

Reinforcement Learning (RL) has emerged as a robust framework for training highly efficient agents to interact with complex environments and learn optimal decision-making policies. RL algorithms aim to devise effective strategies by maximizing cumulative rewards from interactions with the environment. This approach has led to remarkable achievements in diverse fields, including gaming and robotics~\citep{silver2014deterministic,schulman2015high,lillicrap2015continuous,mnih2016asynchronous}. These algorithms often comprise multiple components that are essential for training and adapting neural policies. For example, model-based RL involves learning a model of the world~\citep{racaniere2017imagination,hafner2019dream,janner2019trust,schrittwieser2020mastering} while most model-free policy gradient methods train a value or Q-network to control the variance of the gradient update~\citep{mnih2013playing,schulman2017proximal,haarnoja2018soft,hessel2018rainbow}. Training these multifaceted networks poses challenges due to their nested nature~\citep{boyan1994generalization,anschel2017averaged} and the necessity to extract meaningful features from state-action spaces, coupled with assigning appropriate credit in complex decision-making scenarios. Consequently, these factors contribute to fragile learning procedures, high sensitivity to hyperparameters, and limitations on the network's parameter capacity~\citep{islam2017reproducibility,henderson2018deep,Engstrom2020Implementation}.

To address these challenges, various auxiliary tasks have been proposed, including pre-training different networks to solve various tasks, such as forward or backward dynamics learning~\citep{ha2018world,schwarzer2021pretraining} as well as using online contrastive learning to disentangle feature extraction from task-solving~\citep{laskin2020curl,nachum2021provable,eysenbach2022contrastive}. Alternatively, pre-training agents from a static dataset via offline RL without requiring interaction with the environment also enables robust policies to be deployed for real applications. Most of these approaches rely either on conservative policy optimization~\citep{fujimoto2021minimalist,kumar2020conservative} or supervised training on state-action-rewards trajectory inputs where the transformer architecture has proven to be particularly powerful~\citep{chen2021decision,janner2021offline}.

Recently, self-supervised learning has emerged as a powerful paradigm for pre-training neural networks in various domains including NLP~\citep{chowdhery2022palm,brown2020language,touvron2023llama}, computer vision~\citep{dosovitskiy2020image,bao2021beit,he2022masked} or biology~\citep{lin2023evolutionary,dalla2023nucleotide}, especially when combined with the transformer architecture. Inspired by impressive NLP results with the transformer architecture applied to sequential discrete data, most self-supervised techniques use tokenization, representing input data as a sequence of discrete elements called tokens. Once the data is transformed, first principle objectives such as mask modeling \citep{devlin2018bert} or next token prediction \citep{brown2020language} can be used for self-supervised training of the model. In RL, recent works have explored the use of self-supervised learning to pre-train transformer networks with expert data. While these investigations have yielded exciting outcomes, such as zero-shot capabilities and transfer learning between environments, methods such as \mtm~\citep{wu2023masked} and \smart~\citep{sun2023smart} often rely on highly specific masking techniques and masking schedules~\citep{liu2022masked}, and explore transfer learning across a limited number of tasks. Hence, further exploration of this class of methods is warranted. In this paper, we provide a general study of the different self-supervised objectives and of the different tokenization techniques. In addition, we outline a standardized set of downstream tasks for evaluating the transfer learning performance of pre-trained models, ranging from behavioral cloning to offline RL, robustness to sensor failure, and adaptation to changing dynamics.

\textbf{Our contributions.} With this objective in mind, we introduce the \pasta study, which stands for \underline{p}retrained \underline{a}ction-\underline{s}tate \underline{t}ransformer \underline{a}gents. This study provides comprehensive comparisons that include four pre-training objectives, two tokenization techniques, four pre-training datasets, and 7 downstream tasks. These tasks are categorized into three groups and span four continuous control environments. The \pasta downstream tasks encompass imitation learning and standard RL to demonstrate the versatility of the pre-trained models. In addition, we explore scenarios involving four physical regime changes and 11 observation alterations to assess the robustness of the pre-trained representations. Finally, we assess the zero-shot performance of the models for predictions related to decision-making. We summarize the key findings of our study below:
\begin{enumerate}

\vspace{-0.03cm}\item \textbf{Tokenize trajectories at the component level.} Tokenization at the component level significantly outperforms tokenization at the modality level. In other words, it is more effective to tokenize trajectories based on the individual components of the state and action vectors, rather than directly tokenizing states and actions as is commonly done in existing works.

\vspace{-0.03cm}\item \textbf{Prefer first principle objectives over convoluted ones.} First principle training objectives, such as random masking or next-token prediction with standard hyperparameters match or outperform more intricate and task-specific objectives carefully designed for RL, such as those considered in \mtm or \smart.

\vspace{-0.03cm}\item \textbf{Pre-train the same model on datasets from multiple domains.} Simultaneously pre-training the model on datasets from the four environments leads to enhanced performance across all four environments compared to training separate models for each environment.

\vspace{-0.03cm}\item \textbf{Generalize with a small parameter count.} All of the examined models have fewer than 7 million parameters. Hence, while these approaches are both affordable and practical even on limited hardware resources, the above findings are corroborated by experimentation with four transfer learning scenarios: a) probing (the pre-trained models generate embeddings and only the policy heads are trained to address downstream tasks), b) last layer fine-tuning (only the last layer of the pre-trained model is fine-tuned), c) full fine-tuning and d) zero-shot transfer.

\end{enumerate}

\begin{figure*}[h]
    \centering
    {\includegraphics[width=\linewidth]{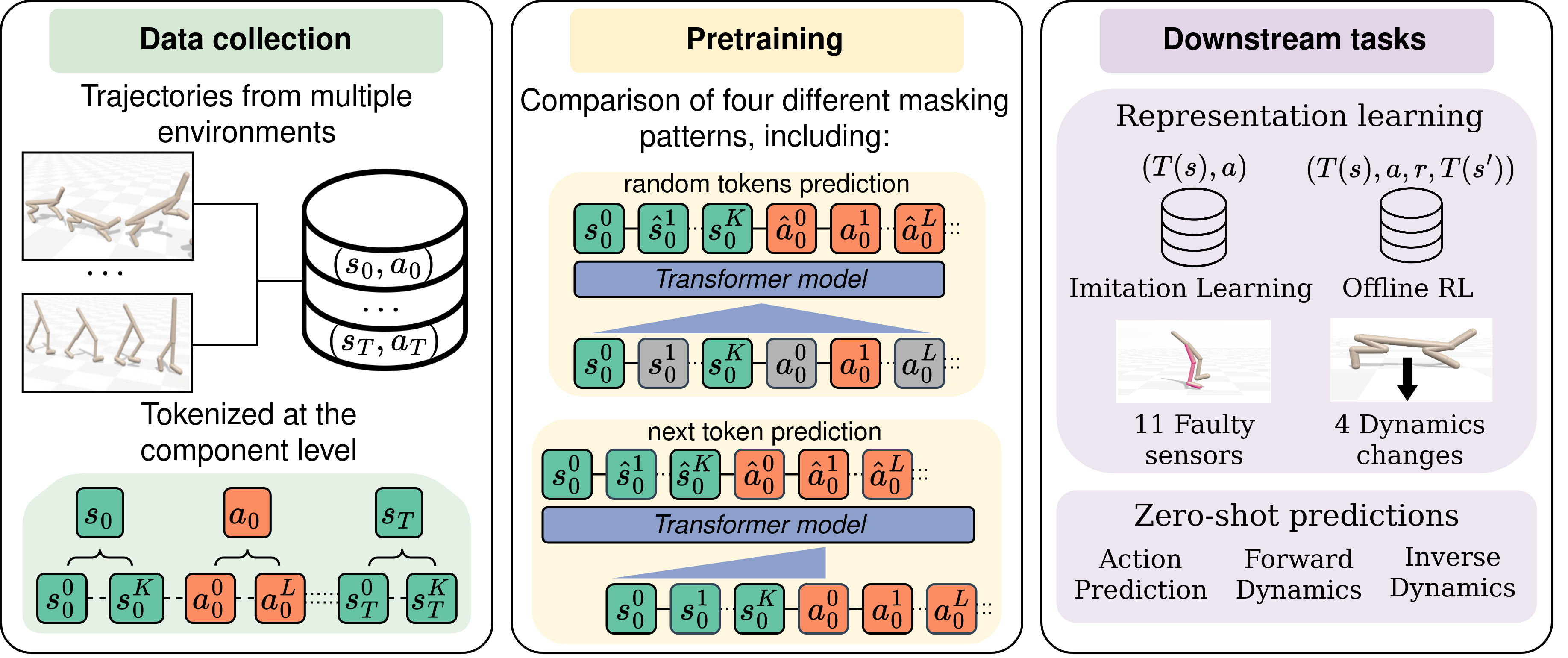}}
    \caption{Illustration of the \pasta study. \textbf{Left:} State-action trajectories are collected from multiple environments and are tokenized at the component level. \textbf{Middle:} A transformer model is pre-trained by processing fixed-size chunks of these sequences. It learns latent representations $T(s)$ of the environments' states. In this study, we compare different tokenization schemes, masking patterns, and pre-training objectives, \eg, random tokens prediction (\bert) or next token prediction (\gpt). \textbf{Right:} The representations of the pre-trained transformer models are evaluated on multiple downstream tasks in which the learned representation $T(s)$ serves as a surrogate state for the policy. Different fine-tuning methods are investigated: probing, last-layer fine-tuning and full fine-tuning.}
    \label{fig:masking}
\end{figure*}

\section{Related Work}
\paragraph{Self-supervised Learning for RL.}

Self-supervised learning, which trains models using unlabeled data, has achieved notable success in various control domains~\citep{liu2021behavior,yuan2022pre,laskin2022cic}. One effective approach is contrastive self-prediction~\citep{chopra2005learning,le2020contrastive,yang2021representation,banino2021coberl} which have proven effective in efficient data augmentation strategies, enabling downstream task solving through fine-tuning, particularly in RL tasks~\citep{laskin2020curl,nachum2021provable}. Our study aligns with this trend, focusing on domain-agnostic self-supervised mechanisms that leverage masked predictions to pre-train general-purpose RL networks.

\paragraph{Offline RL and Imitation Learning.}
Offline learning for control involves leveraging historical data from a fixed behavior policy $\pi_b$ to learn a reward-maximizing policy in an unknown environment. Offline RL methods are typically designed to restrict the learned policy from producing out-of-distribution actions or constrain the learning process within the support of the dataset. Most of these methods usually leverage importance sampling~\citep{sutton2016emphatic,nair2020awac,liu2022offline} or incorporate explicit policy constraints~\citep{kumar2019stabilizing,fujimoto2021minimalist,fakoor2021continuous,dong2023model}. In contrast, Imitation learning (IL) focuses on learning policies by imitating expert demonstrations. Behavior cloning (BC) involves training a policy to mimic expert actions directly while Inverse RL~\citep{ng2000algorithms} aims to infer the underlying reward function to train policies that generalize well to new situations. In contrast, the models investigated in \pasta focus on learning general reward-free representations that can accelerate and facilitate the training of any off-the-shelf offline RL or IL algorithm.

\paragraph{Masked Predictions and Transformers in RL.}
Recently, self-supervised learning techniques based on next token prediction~\citep{brown2020language} and random masked predictions~\citep{devlin2018bert} have gained popularity. These methods involve predicting missing content by masking portions of the input sequence. These first principle pre-training methods have achieved remarkable success in various domains, including NLP~\citep{radford2018improving,radford2019language}, computer vision~\citep{dosovitskiy2020image,bao2021beit,van2017neural}, and robotics~\citep{driess2023palm}. We explore the effectiveness of different variants of these approaches, with various masking patterns and pre-training objectives, in modeling RL trajectories and learning representations of state-action vector components. Transformer networks have been particularly valuable for these purposes. The decision transformer~\citep{chen2021decision} and trajectory transformer~\citep{janner2021offline} have emerged as offline RL approaches using a causal transformer architecture to fit a reward-conditioned policy, paving the way for subsequent work~\citep{zheng2022online,yamagata2022q,liu2022masked,lee2023supervised,badrinath2023waypoint}. Notably, \gato~\citep{reed2022generalist} is a multi-modal behavioral cloning method that directly learns policies, while \pasta focuses on pre-training self-supervised representations. Additionally, \mtm~\citep{wu2023masked} and \smart~\citep{sun2023smart} propose original masking objectives for pre-training transformers in RL. \mtm randomly masks tokens while ensuring some tokens are predicted without future context. It uses modality-level masking and is limited to single-domain pre-training. Conversely, \smart uses a three-fold objective for pre-training a decision transformer with forward-dynamics prediction, inverse-dynamics prediction, and "random masked hindsight control" with a curriculum masking schedule. It focuses on processing real-valued visual observation sequences and investigates generalization across different domains. In \pasta, we compare several first principle pre-training objectives without a masking schedule to these state-of-the-art approaches across multiple environments and diverse downstream tasks.

\section{The \pasta Study}
\label{sec:pasta_study}

\subsection{Preliminaries}

\paragraph{Self-supervised Learning framework.} In this paper, we study self-supervised learning~\citep{balestriero2023cookbook} techniques to pre-train models on a large corpus of static (offline) datasets from interactions with simulated environments, as done in~\citet{shah2021rrl,schwarzer2023bigger}. By solving pre-training objectives, such as predicting future states or filling in missing information, the models learn to extract meaningful features that capture the underlying structure of the data. We focus our study on the use of the transformer architecture due to its ability to model long-range dependencies and capture complex patterns in sequential data.  In addition, the attention mechanism is designed to consider the temporal and intra-modality (position in the state or action vectors) dependencies.
After pre-training the models, we evaluate their capabilities to solve downstream tasks. This analysis is done through the lenses of three mechanisms: (i) probing, (ii) fine-tuning, and (iii) zero-shot transfer. The goal of the study is to investigate which pre-training process makes the model learn the most generalizable representations to provide a strong foundation for adaptation and learning in specified environments. An illustration of the approach adopted in \pasta is given in Figure~\ref{fig:masking}.

\paragraph{Reinforcement Learning framework.} In this paper, we place ourselves in the Markov Decision Processes~\citep{puterman1994} framework. A Markov Decision Process (MDP) is a tuple $M = \{ \mathcal{S}, \mathcal{A}, \mathcal{P}, R, \gamma \}$, where $\mathcal{S}$ is the state space, $\mathcal{A}$ is the action space, $\mathcal{P}$ is the transition kernel, $\mathcal{R}$ is the bounded reward function and $\gamma \in [0, 1)$ is the discount factor. Let $\pi$ denote a stochastic policy mapping states to distributions over actions. We place ourselves in the infinite-horizon setting, \ie, we seek a policy that optimizes $J(\pi) = \mathbb{E}_{\pi}[\sum_{t = 0}^{\infty} \gamma^{t} r\left(s_{t}, a_{t}\right)]$. The value of a state is the quantity $V^{\pi}(s) = \mathbb{E}_{\pi}[\sum_{t = 0}^{\infty} \gamma^{t} r\left(s_{t}, a_{t}\right) | s_0 = s]$ and the value of a state-action pair $Q^\pi (s,a)$ of performing action $a$ in state $s$ and then following policy $\pi$ is defined as: $Q^\pi (s,a) = \mathbb{E}_{\pi}\left[\sum_{t=0}^{\infty} \gamma^{t} r\left(s_{t}, a_{t}\right) | s_0 = s, a_0 = a\right]$.

\subsection{Component-level Tokenization}
A key focus of the \pasta study is the representation of trajectories at the \textit{component-level} for states and actions (\ie, one state component corresponds to one token, as depicted in the middle panel of Figure~\ref{fig:masking}) rather than at the \textit{modality-level} (\ie, one state corresponds to one token). Most previous work, including \smart~\citep{sun2023smart} and \mtm~\citep{wu2023masked} focus on the \textit{modality-level} and consider each trajectory as a sequence of state, action (and often return) tuples, while an alternative is to break the sequences down into individual state and action \textit{components}. Moreover, for the purpose of this study, we exclude the return to focus on general methods applicable to reward-free settings that learn representations not tied to task-specific rewards~\citep{stooke2021decoupling,yarats2021reinforcement}. Based on our experimental results, we argue that the \textit{component-level} level tokenization allows capturing dynamics and dependencies at different space scales, as well as the interplay between the agent's morphological actions and the resulting states. As we observe in Section~\ref{sec:exp}, this results in more generalizable representations that improve the performance of downstream tasks across different robotic structures.

\begin{figure}[t!]
    \centering
    \includegraphics[width=\linewidth]{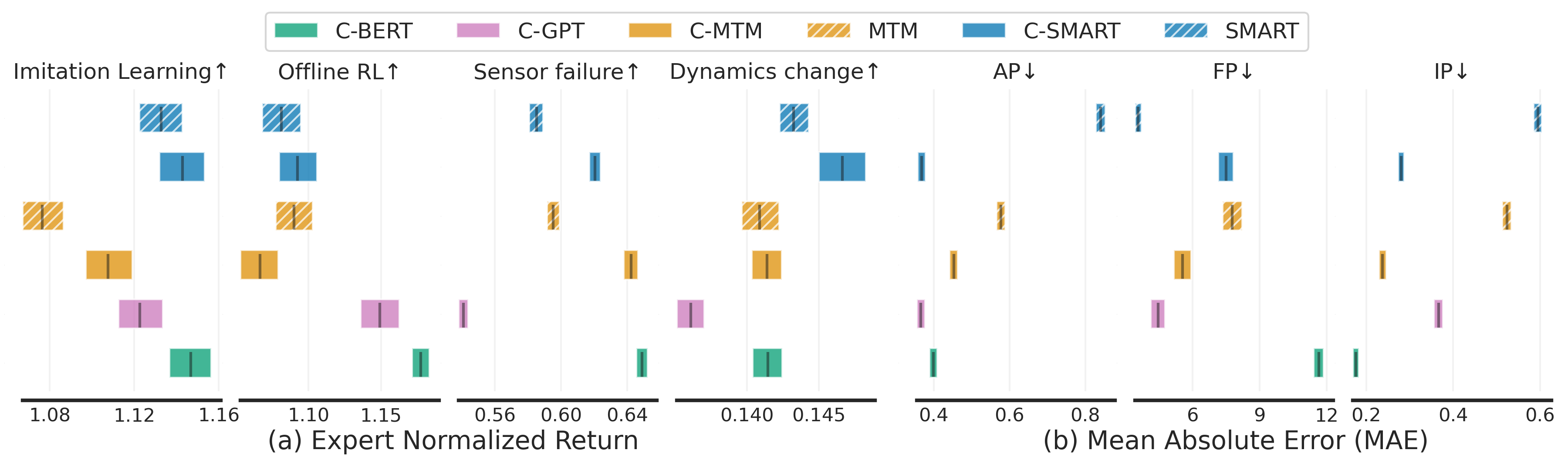}

    \vspace{-10pt}
    \caption{Performance aggregation of the component-level tokenization models (C-*) and modality-level models (\mtm and \smart) with different masking and training objectives. In \textbf{(a)} we report the Interquartile Mean (IQM) of the expert normalized score, computed with stratified bootstrap confidence intervals (CI), obtained in the four fine-tuning downstream tasks over 5 seeds and in \textbf{(b)} the zero-shot transfer tasks: Action Prediction (AP), Forward Prediction (FP), and Inverse Prediction (IP) with 95\% CI. Results are aggregated over all four environments. We developed our own implementation of \mtm and \smart using the same masking patterns and training objectives. ↑ (resp.  ↓) indicates that higher (resp. lower) is better.}
    \label{fig:main_exp}
\end{figure}

\subsection{Pre-training}

\paragraph{Trajectory modeling.}

The \pasta study includes different types of self-supervised learning strategies, each using different combinations of random token masking and/or next token prediction. Next token prediction uses autoregressive masking, while random masked prediction aims to learn from a sequence of trajectory tokens denoted as $\tau = (s^0_0, ..., s^K_0, a^0_0, ..., a^L_0, ..., s^0_T, ..., s^K_T)$. The model's task is to reconstruct this sequence when presented with a masked version $\hat{\tau} = T_\theta (\texttt{Masked}(\tau))$, where K is the observation space size, L is the action space size and T is an arbitrary trajectory size. Here, $T_\theta$ refers to a bi-directional transformer, and $\texttt{Masked}(\tau)$ represents a modified view of $\tau$ where certain elements in the sequence are masked. For instance, a masked view could be $(s^0_0, ..., s^K_0, a^0_0, ..., a^L_0, ..., \_, ..., \_)$, where the underscore “$\_$” symbol denotes a masked element.

\paragraph{Pre-training objectives.}

Next, we introduce the masking patterns investigated in the experimental study. First, the \cgpt masking pattern mimics \gpt's masking mechanism and uses causal (backward-looking) attention to predict the next unseen token in RL trajectories. Second, we have the \cbert masking pattern, derived from \bert's masking mechanism which uses random masks to facilitate diverse learning signals from each trajectory by enabling different combinations. Figure~\ref{fig:masking} provides a visual representation of the \cbert and \cgpt masking mechanisms. Third, the \mtm masking scheme~\citep{wu2023masked} combines random masking (similar to \bert) and causal prediction of the last elements of the trajectory. This latter aims to prevent the model from overly relying on future token information. While \mtm operates at the modality level, we adapt it to operate directly on components by masking random tokens within the trajectory and additionally masking a certain proportion of the last tokens. We refer to this method as \cmtm, which stands for component-level \mtm. Finally, \smart's training objective encompasses three different masking patterns~\citep{sun2023smart}: forward-dynamics, inverse-dynamics and masked hindsight control. The training involves adding up the three losses corresponding to the three masking patterns. Similarly, we derive \csmart, where instead of masking an entire modality at each stage, we mask a random fraction of the tokens within that modality. See Appendix~\ref{ap:smart-mtm} for additional details.

\subsection{Downstream evaluation}
\label{sec:downstream_evaluation}
In this study, we evaluate the effectiveness of \pasta models in transfer learning from two perspectives. Firstly, we examine the ability of pre-trained models to generate high-quality representations. This evaluation is carried out through probing, full fine-tuning and last layer fine-tuning. Secondly, we investigate the capability of pre-trained models to solve new tasks in a zero-shot transfer setting. To accomplish this, we introduce two sets of tasks: \textbf{Representation learning} tasks (4) and \textbf{Zero-shot transfer} tasks (3), comprising a total of 7 evaluation downstream tasks. These task sets are further divided into sub-categories. These categories are designed to provide a general-purpose assessment for pre-trained agents, irrespective of the specific environment or domain.

\begin{figure}[h]
    \centering
    \includegraphics[width=\linewidth]{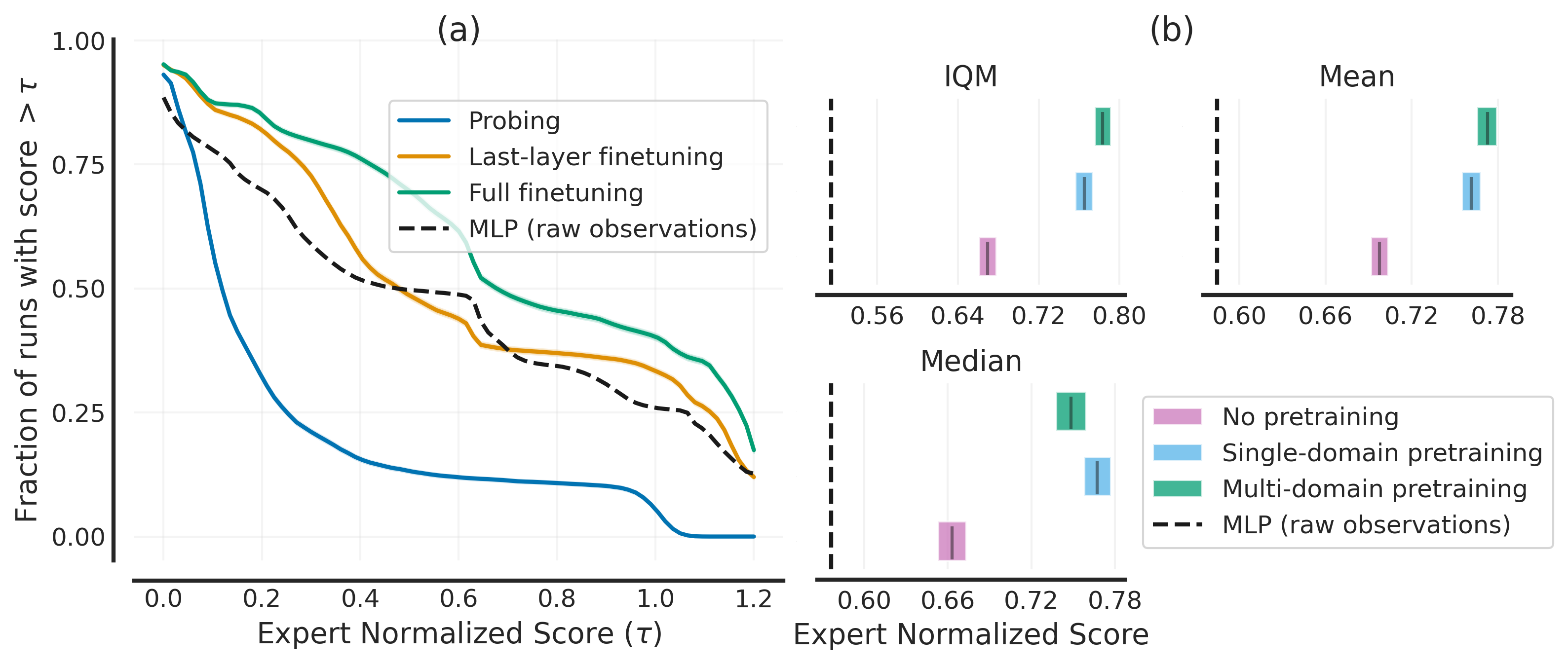}
    \vspace{-15pt}
    \caption{\textbf{(a)} Performance profile of models after full fine-tuning, last layer fine-tuning, no fine-tuning (probing), and RL policies trained from raw observations. Shaded areas show bootstrapped confidence intervals over 5 seeds and 256 rollouts.  \textbf{(b)} Evaluation in all downstream tasks with multi- and single-domain pre-training, no-pretraining and training from raw observations. Remarkably, multi-domain pre-training performs better or on par with single-domain pre-training, despite being trained on the same amount of data.}

    \label{fig:main-results}
\end{figure}

\paragraph{Representation learning.}
The representation learning tasks encompass four sub-categories: Imitation Learning, Offline RL, Sensor Failure, and Dynamics Change. We evaluate the quality of raw representations learned by pre-trained agents using probing on these tasks. In this setting, the weights of the pre-trained models are kept fixed, and the embeddings produced by the final attention layer are fed into a single dense layer network. As the expressive power of such networks is limited, achieving good performance is contingent upon the embeddings containing sufficient information. Furthermore, we assess the quality of the produced representations through full fine-tuning and last layer fine-tuning, where the weights of the pre-trained agents are further updated to solve the downstream tasks. Fine-tuning just the last layer updates only a small fraction of the total weight volume (<1 million parameters), enhancing memory efficiency and lowering the computational resources required.

\paragraph{Zero-shot transfer.}
The zero-shot tasks are organized into three categories: Action Prediction (AP), Forward dynamics Prediction (FP), and Inverse dynamics Prediction (IP). These categories evaluate the pre-trained models' ability to directly predict states or actions based on trajectory information. Specifically, the prediction problems can be expressed as follows; AP: ($\tau_{t-1}, s_t \rightarrow a_t$), FP: ($\tau_{t-1}, s_t, a_t \rightarrow s_{t+1}$) and IP: ($\tau_{t-1}, s_t, s_{t+1} \rightarrow a_{t}$), where the input to the model is shown on the left side of the parentheses, and the prediction target is shown on the right side. For each category, we examine both component prediction and modality (state or action) prediction.

\renewcommand{\arraystretch}{1.1}
\begin{table}[t!]
\centering
\caption{Comparison of agents using representations learned from modality-level tokenization, component-level tokenization, and from an MLP policy network in the four representation learning downstream tasks. We include the maximum performance obtained using modality-level or component-level tokenization. (↑) indicates higher is better and [11] means 11 variations per task. We repeatedly trained all methods with 5 different random seeds and evaluated them using 256 rollouts.}
\label{tab:comparison}
\resizebox{\textwidth}{!}{%
\begin{tabular}{llccc}
\hline
Domain              & \multicolumn{1}{c}{Task} & \begin{tabular}[c]{@{}c@{}}RL policy\\ from raw obs\end{tabular} & \begin{tabular}[c]{@{}c@{}}Modality-level\\ tokenization\end{tabular} & \begin{tabular}[c]{@{}c@{}}Component-level\\ tokenization\end{tabular}  \\ \hline
\multirow{4}{*}{HalfCheetah} & IL (↑) [1]                                & 1.132 ± 0.003                                 & \textbf{1.151 ± 0.003} &  \textbf{1.154 ± 0.003}         \\
                             & Off-RL (↑) [1]                            & 0.571 ± 0.030                                  & \textbf{1.152 ± 0.004}  &  \textbf{1.154 ± 0.003}        \\
                             & Sensor failure (↑) [11]                    & 0.896 ± 0.003                                  & 1.006 ± 0.002  &  \textbf{1.048   ±  0.002}     \\
                             & Dynamics change (↑) [4]                   & 0.251 ± 0.003                                  & 0.339 ± 0.003  &  \textbf{0.369   ± 0.004}        \\ \hline
\multirow{4}{*}{Hopper}      & IL (↑) [1]                                & 0.898 ± 0.022                                    & 0.847 ± 0.019  &   \textbf{1.078 ± 0.021}     \\
                             & Off-RL (↑) [1]                            & 0.890 ± 0.022                                   & 0.812 ± 0.020  &  \textbf{0.971 ± 0.022}        \\
                             & Sensor failure (↑) [11]                    & 0.307 ± 0.005                                   & 0.554 ± 0.006  &   \textbf{0.584  ± 0.007}       \\
                             & Dynamics change (↑) [4]                   & 0.169 ± 0.035                                   & \textbf{0.290 ± 0.035}  &   \textbf{0.290  ±  0.038}     \\ \hline
\multirow{4}{*}{Walker2d}    & IL (↑) [1]                                & 0.736 ± 0.010                                   & 1.128 ± 0.029   &   \textbf{1.178 ± 0.031}     \\
                             & Off-RL (↑) [1]                            & 0.911 ± 0.025                                  & 0.923 ± 0.025   &  \textbf{1.046  ±  0.023}     \\
                             & Sensor failure (↑) [11]                    & 0.339 ± 0.003                                  & 0.419 ± 0.003  &  \textbf{0.511 ± 0.003}      \\
                             & Dynamics change (↑) [4]                  & 0.000 ± 0.000                                     & \textbf{0.004 ± 0.001}    &   \textbf{0.005 ± 0.001}     \\ \hline
\multirow{4}{*}{Ant}    & IL (↑) [1]                                & 0.876 ± 0.032                                     & \textbf{1.203 ± 0.008}   &   \textbf{1.209 ± 0.005}      \\
                             & Off-RL (↑) [1]                            & 0.846 ± 0.030                                  & 0.907 ± 0.035   &  \textbf{1.213 ± 0.021}      \\
                             & Sensor failure (↑) [11]                    & 0.082 ± 0.004                                  & 0.615 ± 0.007  &  \textbf{0.717   ±  0.007}      \\
                             & Dynamics change (↑) [4]                  & 0.015 ± 0.001                                        & 0.065 ± 0.001    &   \textbf{0.068 ± 0.001}     \\ \hline
\end{tabular}%
}

\end{table}

\section{Experimental Analysis}
\label{sec:exp}

In this section, we present the experimental study conducted to examine the impact of pre-training objectives, tokenization, and dataset preparation choices on the generalization capabilities of pre-trained \pasta models.

\subsection{Experimental Setup}

\paragraph{Domains.}
To assess the effectiveness of our approach, we select tasks from the Brax library~\citep{brax2021github}, which provides environments designed to closely match~\citep{freeman2021brax} the original versions found in MuJoCo's environment suite~\citep{todorov2012mujoco}. Brax provides significant advantages over MuJoCo, as it offers a highly flexible and scalable framework for simulating robotic systems based on realistic physics. More information about the environments is given in Appendix~\ref{ap:envs}. The pre-training datasets consist of trajectories collected from four Brax environments: HalfCheetah, Hopper, Walker2d and Ant. Following the protocols used in previous work~\citep{fu2020d4rl,sun2023smart}, we trained 10 Soft Actor-Critic (SAC)~\citep{haarnoja2018soft} agents initialized with different seeds and collected single- and multi-domain datasets composed of 680 million tokens in total. For details about the pre-training datasets, we refer the reader to Appendix~\ref{ap:dataset}.

Consequently, the 7 downstream tasks presented in Section~\ref{sec:pasta_study} are set up for each environment resulting in a total of 28 tasks across environments. The reason we introduce multiple environments for pre-training and evaluation is (i) to evaluate the reproducibility of our findings across domains and (ii) to study the performance of models pre-trained on the four datasets simultaneously (multi-domains model) compared to specialized single-domain models. For further details about the implementation of downstream tasks, please refer to Appendix~\ref{sec:downstream}.

\paragraph{Implementation details.}
In this study, we focus on reasonably sized and efficient models, typically consisting of around 7 million parameters. To capture positional information effectively, we incorporate a learned positional embedding layer at the component level. Additionally, we include a rotary position encoding layer following the approach in~\citet{su2021roformer} to account for relative positional information. More implementation details are provided in Appendix~\ref{ap:implem}. To convert the collected data (state or action components) into tokens, we adopt a tokenization scheme similar to~\citet{reed2022generalist}. Continuous values are mu-law encoded to the range [-1, 1] and discretized into 1024 uniform bins. The sequence order follows observation tokens followed by action tokens, with transitions arranged in timestep order.

\paragraph{Baselines}
To put in perspective the performance attained by the different pre-trained models, we compare them with a simple MLP architecture, that learns directly from the raw observations with no pre-training. For fairness, the architecture and parameters of the MLP have been tuned by taking the best performance independently on each domain and on each downstream task.

\subsection{Results}
\paragraph{Component-level Tokenization.}

Our initial analysis probes the influence of tokenization detail -- how finely we dissect the data -- on the models' ability to generalize. We scrutinize this by training models using both the \smart and \mtm protocols at two levels of granularity: modality-level (predicting entire observations and actions) for \smart and \mtm, and component-level (focusing on individual observation and action elements) for \csmart and \cmtm. Despite sharing identical architectures and training conditions, and being trained on the same multi-domain dataset, the models' fine-tuning performances vary. As depicted in Figure~\ref{fig:main_exp} (a), a shift from modality-level to component-level tokenization markedly enhances model performance across a spectrum of tasks, including Imitation Learning, Offline RL, variations of Sensor Failure, and Dynamics Change tasks. Furthermore, Table~\ref{tab:comparison} provides a breakdown of performance for both tokenization techniques across different domains. Overall, we observe that transitioning from modality-level to component-level tokenization improves performance.

\paragraph{Masking objectives.}
Subsequently, we compare fundamental tokenization approaches, \ie, masked language modeling (\bert) and next token prediction (\gpt) against state-of-the-art transformer RL methods \mtm and \smart which incorporate more tailored design choices. In the light of the previous section demonstrating the advantages of using component-level tokenization, we design \cbert and \cgpt which focus on individual observation and action elements. These models are trained under similar conditions as \csmart and \cmtm on the multi-domain dataset. We systematically fine-tune all models for all downstream tasks and domains. Figure~\ref{fig:main_exp} (a) reveals that \cbert exhibits on average higher performance on the considered downstream tasks compared to other masking schemes and training objectives. This demonstrates that simpler objectives are sufficient to achieve robust generalization performance. Based on \cbert showing the best performance among other models, it is selected for further analysis within this study.

\paragraph{Multi-domain representation.}
We now investigate the benefits of pre-training models on multi-domain representations, using a granular, component-level tokenization approach. Figure~\ref{fig:main-results} (b) presents an aggregated comparison of multi-domain and single-domain representation models across all tasks and domains, with each model being trained on the same number of tokens. We compare these approaches with a randomly initialized model and our MLP policy network baseline. Our findings confirm that policies trained with multi-domain representations using component-level tokenization outperform policies trained from raw observations with neural networks comprised of an equivalent number of parameters (cf. Appendix~\ref{sec:fair}). This validates the capability of the model to produce useful representations. We also compare the performance of multi-domain models against a randomly initialized model (no pre-training) with the same architecture and confirm the positive effect of pre-training on the observed performance. Then, comparing multi-domain models against specialized models trained in isolation for each environment, we show that specialized models are slightly outperformed in terms of final performance, suggesting that multi-domain models can compress the information contained in the representation of the four domains into one single model by sharing most of the parameters between the different domains and without performance degradation. Note that to ensure a fair comparison, all models were trained on the same amount of tokens and have the same representation capability (architecture and number of learned parameters). For a detailed breakdown of the results for each downstream task, please refer to panels provided in Appendix~\ref{sec:breakdown-results}.

\paragraph{Fine-tuning.}
Figure~\ref{fig:main-results} (a) presents the performance profiles for various fine-tuning strategies: probing (the transformer's parameters are frozen), last layer fine-tuning (only the last layer's parameters are trained) and full fine-tuning (all the parameters are trainable). We observe that full fine-tuning results in a higher fraction of runs achieving near-expert scores, followed by last-layer fine-tuning, MLP, and probing. We note that the performance of the probed model may initially seem unexpected, yet several underlying factors could contribute to this outcome. Primarily, it suggests that the pre-training tasks may not align closely with the downstream tasks, leading to representations that, although rich, are not directly applicable. This mirrors that of LLMs which require task-specific fine-tuning for particular NLP applications. The results show that the fine-tuning process appears to bridge the gap between the generic representations and the specialized requirements of the downstream tasks. 

\paragraph{Robust representations.}
\label{sec:robust}
\begin{table}[t]
\caption{Breakdown of Expert-normalized returns in the Sensor Failure and Dynamics Change tasks. (↑) indicates that higher is better.}\label{tab:sensor-gravity}
\centering
\footnotesize
    \begin{tabular}{lcc}
    \toprule
    \textbf{Model} & \textbf{\makecell{Sensor Failure}} & \textbf{\makecell{Dynamics Change}} \\
    \midrule
    Multi-domain pre-training (↑) & $0.69 \pm 0.01$ & $0.17 \pm 0.01$  \\
    Single-domain pre-training (↑) & $0.66 \pm 0.01$ & $0.18 \pm 0.01$ \\
    MLP (raw observations) (↑) & $0.41 \pm 0.02$ & $0.11 \pm 0.01$ \\
    No pre-training (↑) & $0.55 \pm 0.01$ & $0.16 \pm 0.01$ \\
    \bottomrule
  \end{tabular}

\end{table}

In this section, we focus on resilience to sensor failure and adaptability to dynamics change. These factors play a crucial role in real-world robotics scenarios, where sensor malfunctions and environmental variations can pose risks and impact decision-making processes. We used BC as the training algorithm and during evaluation, we systematically disabled each of the 11 sensors individually by assigning a value of 0 to the corresponding coordinate in the state vector. In Table~\ref{tab:sensor-gravity}, multi-domain models exhibit higher performance compared to the baselines, demonstrating their enhanced robustness in handling sensor failures. Furthermore, we introduced four gravity changes during the inference phase, and the results reaffirm the resilience of multi-domain learning in adapting to dynamics change, corroborating our previous findings.

\paragraph{Zero-shot predictions.}
In this section, we extend our study into the zero-shot capabilities of pre-trained models. We evaluated an additional suite of tasks, outlined in Section~\ref{sec:downstream_evaluation}, originally introduced in MTM~\citep{he2022masked}. Notably, Figure~\ref{fig:main_exp} (b) reveals that the errors in Action Prediction (AP), Forward Prediction (FP), and Inverse Prediction (IP) for \cgpt and \cbert are on par with those of more sophisticated models like \cmtm or \csmart. This suggests that even simple pre-training objectives are well-aligned with the inference tasks, despite the models not being explicitly trained for these tasks. Such findings reinforce the effectiveness of simple objective functions combined with straightforward masking patterns and component-level tokenization. Importantly, we note that the masking strategy of \cbert and \cgpt allows the emergence of competitive Action Prediction (AP) performance, which, according to the results in Figure~\ref{fig:main_exp} (a) is sufficient and indicative of strong downstream performance.

\section{Discussion}

This paper introduces the \pasta study on Pretrained Action-State Transformer Agents, aiming to deeply investigate self-supervised learning models in reinforcement learning (RL). The study contributes datasets, seven downstream tasks, and analyses across four training objectives and two tokenization methods. Conducted in four different continuous control environments, it showcases the flexibility and efficacy of pre-trained models in transfer learning scenarios, including probing, fine-tuning, and zero-shot evaluation.

One key finding is the effectiveness of fundamental self-supervised objectives over more complex ones. Simple tasks like random masking or next token prediction were as effective, if not more, than bespoke RL objectives. Component-level tokenization also proved superior to modality-level, underscoring the importance of nuanced tokenization for richer representations. Additionally, pre-training on multi-domain datasets led to better performance than domain-specific models, demonstrating the value of cross-domain knowledge transfer. Additionally, the investigation in Section~\ref{sec:robust} highlighted the importance of developing such models that can effectively adapt and make decisions in the presence of sensor failures or dynamic changes, ensuring safety and mitigating risks in robotics applications.

Overall, the findings from this study provide valuable guidance to researchers interested in leveraging self-supervised learning to improve RL in complex decision-making tasks. The models presented in this study are relatively lightweight, enabling the replication of both pre-training and fine-tuning experiments on readily available hardware. In future work, we anticipate further exploration of self-supervised objectives, tokenization methods, and a broader spectrum of tasks to evaluate adaptability and robustness, enhancing the practical applicability of pre-trained agents in real-world scenarios.

\section{Acknowledgments}
Research supported with Cloud TPUs from Google's TPU Research Cloud (TRC).

\onecolumn
\bibliographystyle{iclr2023_conference}
\bibliography{biblio}

\clearpage
\appendix

\section{Additional Experiments}
\subsection{Detailed Breakdown of Downstream Tasks Results}
\label{sec:breakdown-results}

\begin{figure*}[h]
    \centering
    {\includegraphics[width=\linewidth]{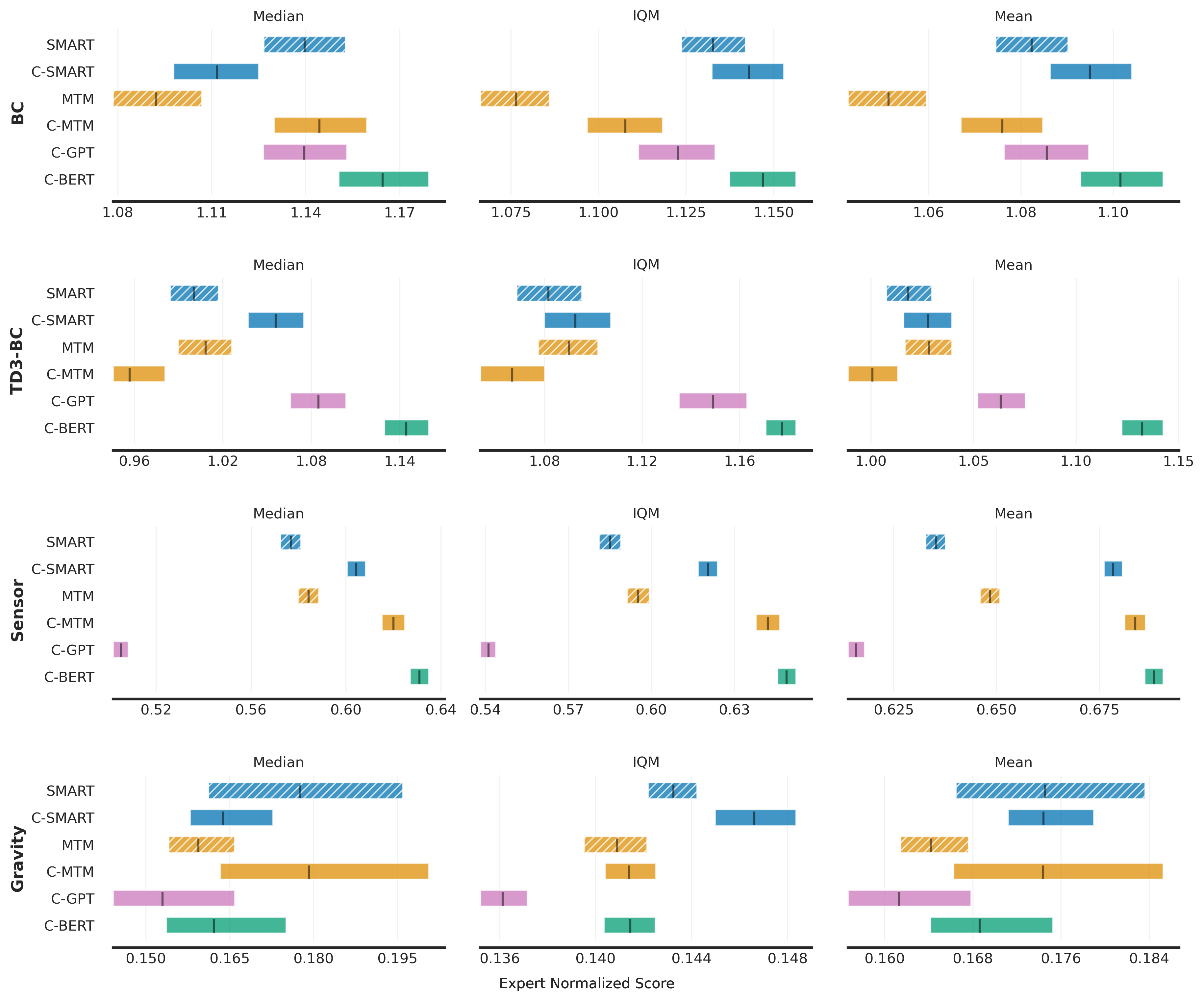}}

    \caption{Detailed breakdown of the Mean, Interquartile Mean (IQM) and Median expert normalized scores, computed with stratified bootstrap confidence intervals, obtained in the four fine-tuning downstream tasks for the four environments HalfCheetah, Hopper, Walker2d and Ant. We repeatedly trained all methods with 5 different random seeds and evaluated them using 256 rollouts.}
    \label{fig:panel_gravity_il_offrl}
\end{figure*}

\section{Implementation Details}
\label{ap:implem}
In the sequence tokenization phase, we do not use return conditioning but since the representation models are pre-trained on multiple environments and tasks, we use environment conditioning, \ie, during training, an environment token is appended at the beginning of the sequences in each batch, providing the model with additional contextual information. In practice, the length of the last two modalities (state and action concatenated) varies across different environments. Therefore, the maximum portion of masked tokens at the end of the sequence differs depending on the environment. For instance, in the Hopper environment with 3 actions and 11 observation tokens, the maximum portion of masked tokens is 14, while in HalfCheetah with 6 actions and 18 observation tokens, it is 24. Additionally, as we maintain a fixed-size context window of 128, the sequences' starting points will have varying truncations for different environments, ensuring a non-truncated state at the end of the window. Another design choice is the embedding aggregation, \ie, how to come from a context\_window x embedding\_dimension tensor to a 1 x embedding\_dimension tensor. We decided to use take the embedding from the last observed token.

\paragraph{Computational Cost.} A significant advantage of the component-level sequencing approach is its reduced input dimension, allowing cheaper computational costs. By capturing the components of states and actions at different time steps, the input space expands linearly rather than quadratically mitigating the challenges associated with the curse of dimensionality. To illustrate this, consider a simple example of a 2-dimensional state space with a discretization size of 9. With a component-level granularity, the input size becomes $2 \times 9 = 18$. In contrast, a state-level granularity results in an input size of $9 \times 9 = 81$. The former exhibits linear growth within the observation space, while the latter demonstrates quadratic growth. Moreover, while it effectively multiplies the length of the input sequence by the average number of components in a state, this drawback is absorbed by the increased context window of transformer models. Lastly, for an equal number of trajectories, the number of tokens is also trivially larger than that with a state- and action-level granularity.

\section{Additional Details on Masking Patterns}
\label{ap:smart-mtm}

In this section, we provide further details on the masking patterns and schedule used in the SMART~\citep{sun2023smart} and MTM~\citep{wu2023masked} baselines. In \cgpt or \cbert, we focused on reducing the technicalities to their minimum: a simple masking pattern, \ie, GPT-like or BERT-like, and no masking schedule.

In SMART, the objective involves three components: Forward Dynamics Prediction, Inverse Dynamics Prediction, and Random Masked Hindsight Control. The masking schedule involves two masking sizes, $k$ and $k'$, which determine the number of masked actions and observations during pre-training. The masking schedule for actions ($k$) is designed to gradually increase the difficulty of the random masked hindsight control task. It starts with $k = 1$, ensuring the model initially predicts masked actions based on a single observed action. As training progresses, the value of $k$ is increased in a curriculum fashion. The masking schedule for observations ($k'$) ensures that the model learns to predict masked actions based on a revealed subsequence of observations and actions, rather than relying solely on local dynamics. Similar to the action masking schedule, $k'$ starts at $1$ and gradually increases during training. SMART's paper suggests that the masking schedule is essential for effective pre-training in control environments. By gradually increasing the masking difficulty, the model is exposed to a range of training scenarios, starting with simple local dynamics and gradually transitioning to complex long-term dependencies.

In MTM, the masking pattern is implemented by requiring at least one token in the masked sequence to be autoregressive, which means it must be predicted based solely on previous tokens, and all future tokens are masked. In addition, MTM uses a modality-specific encoder to elevate the raw trajectory inputs to a common representation space for the tokens. Finally, MTM is trained with a range (between 0.0 and 0.6) of randomly sampled masking ratios.

\section{Experimental Details and Hyperparameters}
In this section, we provide more details about the experiments, including hyperparameter configuration and details of each environment (\eg, version). For all experiments, we run 256 rollouts with 5 different random seeds and report the mean and stratified bootstrap confidence intervals.

\subsection{Fair Comparison}
\label{sec:fair}
To ensure a fair comparison between the representation models using an MLP or a transformer architecture, we made sure to have a comparable number of parameters. Both models consist of a minimum of three layers with a size of 256 for the baseline, while transformer models use a single layer with a hidden size of 512 for the policy. We tested bigger architecture for the MLP without performance gain.

Moreover, we choose to fine-tune the MLP baselines to achieve the best performance in each environment. In contrast, we use the same set of hyperparameters for all domains involving \pasta models. This approach puts \pasta models at a slight disadvantage while holding the promise of potentially achieving even better performance with the \pasta methods with further hyperparameter tuning.

Finally, when a pre-trained model is involved, we always select the final checkpoint after the fixed 3 epochs done over the pre-training dataset.

\subsection{Environment Details}
\label{ap:envs}

\begin{figure*}[!ht]
    \centering
    \subfloat{{\includegraphics[width=.24\linewidth]{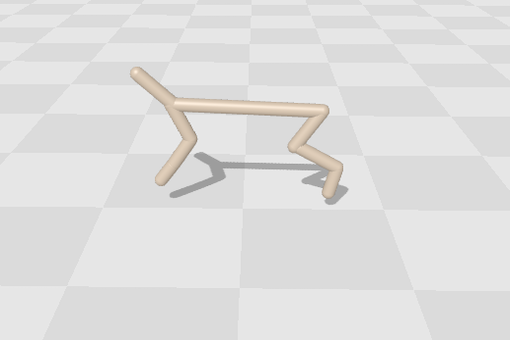}}\quad{\includegraphics[width=.24\linewidth]{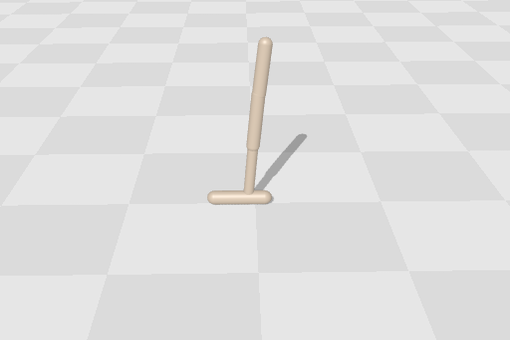}}\quad{\includegraphics[width=.24\linewidth]{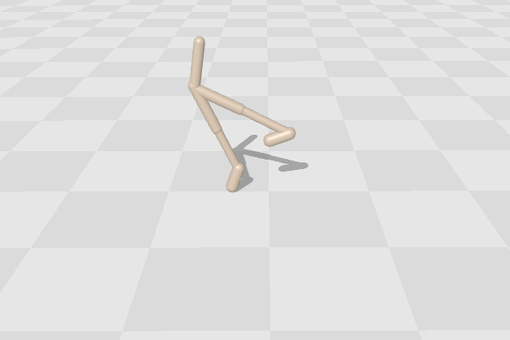}}\quad{\includegraphics[width=.24\linewidth]{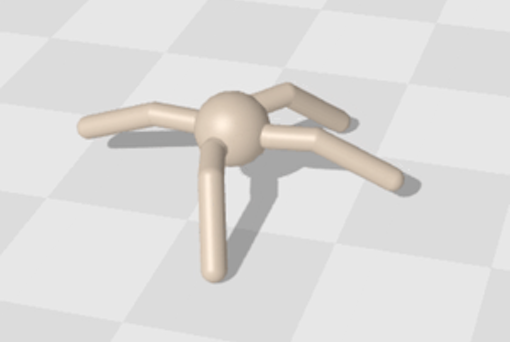}}}
    \caption{Continuous Control Downstream Tasks.}
    \label{fig:envs}
\end{figure*}

For all experiments, we use the $\mathrm{0.0.15}$ version of Brax~\citep{brax2021github}. Each environment in Brax, illustrated in Figure~\ref{fig:envs}, provides a realistic physics simulation, enabling agents to interact with objects and the environment in a physically plausible manner. The tasks studied in this paper feature (i) a HalfCheetah robot~\citep{wawrzynski2009cat} with 9 links and 8 joints. The objective is to apply torques on the joints to make the cheetah run forward as fast as possible. The action space for the agents consists of a 6-element vector representing torques applied between the different links; (ii) a Hopper robot~\citep{erez2011infinite} which is a two-dimensional one-legged figure consisting of four main body parts: the torso, thigh, leg, and foot. The objective is to make hops in the forward direction by applying torques on the hinges connecting the body parts. The action space for the agent is a 3-element vector representing the torques applied to the thigh, leg, and foot joints; (iii) a Walker robot~\citep{erez2011infinite} which is a two-dimensional two-legged figure comprising a single torso at the top, two thighs below the torso, two legs below the thighs, and two feet attached to the legs. The objective is to coordinate the movements of both sets of feet, legs, and thighs to achieve forward motion in the right direction. The action space for the agent is a 6-element vector representing the torques applied to the thigh, leg, foot, left thigh, left leg, and left foot joints; (iv) an Ant robot~\citep{schulman2015high} which is a one torso body with four legs attached to it with each leg having two body parts. The objective is to coordinate the movements of the four legs to achieve forward motion in the right direction. The action space for the agent is an 8-element vector representing the torques applied at the hinge joints.

\subsection{Dataset Details}
\label{ap:dataset}
In this section, we provide further detail on the collection of the datasets.  We trained 10 SAC~\citep{haarnoja2018soft} agents for a total of 5 million timesteps in each of the four environments. From each, we select the 20\% latest trajectories of size 1000, resulting in a combined total of 40 million transitions. With each environment comprising different observation and action sizes, the overall multi-domain dataset is composed of 680 million tokens. We also have one dataset for each domain.

Next, we give the hyperparameters of the SAC agents used to collect the pre-training trajectories. These are given in Table~\ref{tab:hp-sac}.

\setlength{\tabcolsep}{4pt}
\begin{table}[!h]
    \centering
    \caption{Hyperparameters used in SAC.} \label{tab:hp-sac}
    \vspace{2pt}
    \begin{tabular}{lc}
      \toprule 
      \bfseries Hyperparameter & \bfseries Value\\
            \midrule 
Adam stepsize                         & $3 \cdot 10^{-4}$ \\
Discount ($\gamma$)                   & 0.99              \\
Replay buffer size                    & $10^6$              \\
Batch size                            & 256                 \\
Nb. hidden layers                     & 2                 \\
Nb. hidden units per layer            & 256                  \\
Nonlinearity                          & ReLU                \\
Target smoothing coefficient ($\tau$) & 0.005            \\
Target update interval                & 1               \\
Gradient steps per timestep                        & 1           \\
Training steps & 20,000 \\
      \bottomrule 
    \end{tabular}
\end{table}

We also provide a concrete example of the state and action components with their corresponding properties for the simplest robot structure, Hopper. The number of components for each property is given in parentheses. In this case, the action space consists of torques applied to the rotors (3), while the observation space includes the following components: z-coordinate of the top (1), angle (4), velocity (2), and angular velocity (4).

\subsection{Downstream Tasks Details}
\label{sec:downstream}
In this section, we provide the hyperparameters used in the training of the imitation learning algorithm Behavioural Cloning (BC) (Table~\ref{tab:hyper-bc}) and the offline RL algorithm TD3-BC (Table~\ref{tab:hyper-td3bc}).
\begin{table}[h]
\centering
\setlength{\tabcolsep}{8pt}
\caption{Hyperparameters used in the BC downstream task.}
\begin{tabular}{l | l}
Parameter                             & Value             \\ \hline
\addlinespace[0.5ex]
Horizon $T$                       & 1000              \\
Batch Size                      & 1024              \\
Non-Linearity                          & GELU~\citep{hendrycks2016gaussian} \\
Nb. hidden layers                     & 1                 \\
Nb. hidden units per layer            & 512                  \\
Adam stepsize                         & $3 \cdot 10^{-4}$ \\
Training steps & 80,000 \\
\end{tabular}
  \label{tab:hyper-bc}
\end{table}

\begin{table}[h]
\centering
\setlength{\tabcolsep}{8pt}
\caption{Hyperparameters used in the TD3-BC downstream task.}
\begin{tabular}{l | l}
Parameter                             & Value             \\ \hline
\addlinespace[0.5ex]
Horizon $T$                       & 1000              \\
Batch Size                      & 1024              \\
Discount $\gamma$                   & 0.99              \\
Non-Linearity                          & GELU~\citep{hendrycks2016gaussian} \\
Nb. hidden layers                     & 1                 \\
Nb. hidden units per layer            & 512                  \\
Adam stepsize (actor)                         & $1 \cdot 10^{-4}$ \\
Adam stepsize (critic)                         & $3 \cdot 10^{-4}$ \\
Target update rate & $5 \cdot 10^{-3}$ \\
Policy noise &  0.2 \\
Policy noise clipping &  (-0.5, 0.5) \\
Policy update frequency &  2 \\
Conservatism coefficient $\alpha$ & 2.5  \\
Training steps & 140,000 \\
\end{tabular}
  \label{tab:hyper-td3bc}
\end{table}

Then, we give additional details about the Sensor Failures downstream task. In Table~\ref{tab:sensors-hc},~\ref{tab:sensors-hopper},~\ref{tab:sensors-walker} and~\ref{tab:sensors-ant} we include the correspondence between each sensor number and its associated name in all environments. In the 11 variations of the Sensor Failure downstream task, we switch off each one of these sensors.

\setlength{\tabcolsep}{4pt}
\begin{table}[!h]
    \centering
    \caption{Sensor name / Sensor number in Halfcheetah.} \label{tab:sensors-hc}
    \vspace{2pt}
    \begin{tabular}{lc}
      \toprule 
      \bfseries Sensor name & \bfseries Sensor number\\
      \midrule 
      z-coordinate of the center of mass & 1\\
      w-orientation of the front tip & 2\\
      y-orientation of the front tip  & 3 \\
      angle of the back thigh rotor  & 4 \\
      angle of the back shin rotor         & 5 \\
      angle of the back foot rotor         & 6 \\
      velocity of the tip along the y-axis & 7 \\
      angular velocity of front tip        & 8 \\
      angular velocity of second rotor     & 9 \\
      x-coordinate of the front tip        & 10 \\
      y-coordinate of the front tip        & 11 \\
      \bottomrule 
    \end{tabular}
\end{table}

\setlength{\tabcolsep}{4pt}
\begin{table}[!h]
    \centering
    \caption{Sensor name / Sensor number in Hopper.} \label{tab:sensors-hopper}
    \vspace{2pt}
    \begin{tabular}{lc}
      \toprule 
      \bfseries Sensor name & \bfseries Sensor number\\
      \midrule 
      z-coordinate of the top (height of hopper)       & 1\\
      angle of the top                                & 2\\
      angle of the thigh joint                          & 3 \\
      angle of the leg joint                            & 4 \\
      angle of the foot joint                                  & 5 \\
      velocity of the x-coordinate of the top                  & 6 \\
      velocity of the z-coordinate (height) of the top & 7 \\
      angular velocity of the angle of the top                & 8 \\
      angular velocity of the thigh hinge                  & 9 \\
      angular velocity of the leg hinge                       & 10 \\
      angular velocity of the foot hinge                      & 11 \\
      \bottomrule 
    \end{tabular}
\end{table}

\setlength{\tabcolsep}{4pt}
\begin{table}[!h]
    \centering
    \caption{Sensor name / Sensor number in Walker2d.} \label{tab:sensors-walker}
    \vspace{2pt}
    \begin{tabular}{lc}
      \toprule 
      \bfseries Sensor name & \bfseries Sensor number\\
      \midrule 
      z-coordinate of the top (height of hopper)       & 1\\
      angle of the top                                 & 2\\
      angle of the thigh joint                          & 3 \\
      angle of the leg joint                            & 4 \\
      angle of the foot joint                                  & 5 \\
      angle of the left thigh joint                            & 6 \\
      angle of the left leg joint                      & 7 \\
      angle of the left foot joint                            & 8 \\
      velocity of the x-coordinate of the top              & 9 \\
      velocity of the z-coordinate (height) of the top        & 10 \\
      angular velocity of the angle of the top                & 11 \\
      \bottomrule 
    \end{tabular}
\end{table}

\setlength{\tabcolsep}{4pt}
\begin{table}[!h]
    \centering
    \caption{Sensor name / Sensor number in Ant.} \label{tab:sensors-ant}
    \vspace{2pt}
    \begin{tabular}{lc}
      \toprule 
      \bfseries Sensor name & \bfseries Sensor number\\
      \midrule 
      z-coordinate of the torso (centre)       & 1\\
      x-orientation of the torso (centre)                                 & 2\\
      y-orientation of the torso (centre)                        & 3 \\
      z-orientation of the torso (centre)                            & 4 \\
      w-orientation of the torso (centre)                                & 5 \\
      angle between torso and first link on front left                         & 6 \\
      angle between the two links on the front left                    & 7 \\
      angle between torso and first link on front right                            & 8 \\
      angle between the two links on the front right            & 9 \\
      angle between torso and first link on back left        & 10 \\
      angle between the two links on the back left               & 11 \\
      \bottomrule 
    \end{tabular}
\end{table}

Finally, to implement the Dynamics Change downstream task we use the GravityWrapper for Brax environments of the QDax library~\citep{chalumeau2023qdax} and similarly to ~\citet{chalumeau2022neuroevolution} we train the policies with a gravity multiplier of $1$ and we vary this coefficient at inference by the following constant values: $0.1$, $0.25$, $4$, and $10$.


\newpage
\subsection{Hyperparameters}
In Table~\ref{tab:hp}, we show the hyperparameter configuration for \cbert across all experiments.

\setlength{\tabcolsep}{4pt}
\begin{table}[!h]
    \centering
    \caption{Hyperparameters and configuration details for \cbert across all experiments.} \label{tab:hp}
    \vspace{2pt}
    \begin{tabular}{lc}
      \toprule 
      \bfseries Hyperparameter & \bfseries Value\\
      \midrule 
      Transformer Layers & 10\\
      Transformer Heads & 8\\
      Noising Ratio & 0.15 \\
      Masking Probability & 0.8 \\
      Random Token Probability & 0.1 \\
      Non-Linearity & GELU \\
      Learning Rate & $3e-4$ \\
      Num Epochs & 3 \\
      Batch Size & 4096 \\
      Num Quantization Tokens & 1024 \\
      Embedding Dimension & 256 \\
      \bottomrule 
    \end{tabular}
\end{table}

\end{document}